\documentclass[conference, a4paper]{IEEEtran}
\IEEEoverridecommandlockouts
\usepackage{mathtools}
\usepackage{amssymb,amsmath}
\usepackage{ifpdf}
\usepackage{url}
\usepackage{graphicx}
\usepackage{caption}
\usepackage{subcaption}
\usepackage{cite}
\usepackage{etoolbox}
\usepackage{multirow}
\usepackage{tabu}
\usepackage{wrapfig}
\newtheorem{definition}{Definition}
\usepackage{array}
\usepackage{hyperref}
\newcolumntype{L}[1]{>{\raggedright\let\newline\\\arraybackslash\hspace{0pt}}m{#1}}
\newcolumntype{C}[1]{>{\centering\let\newline\\\arraybackslash\hspace{0pt}}m{#1}}
\newcolumntype{R}[1]{>{\raggedleft\let\newline\\\arraybackslash\hspace{0pt}}m{#1}}

\begin{document}

\title{Reduced Robust Random Cut Forest for Out-Of-Distribution detection in machine learning models }

\author{\IEEEauthorblockN{Harsh Vardhan}
\IEEEauthorblockA{\textit{Institute for Software and Integrated System} \\
\textit{Vanderbilt University}\\
Nashville, USA \\
harsh.vardhan@vanderbilt.edu}
\and
\IEEEauthorblockN{Janos Sztipanovits}
\IEEEauthorblockA{\textit{Institute for Software and Integrated System} \\
\textit{Vanderbilt University}\\
Nashville, USA \\
janos.szpitanovits@vanderbilt.edu}
}

\maketitle
\subsection*{\textbf{\textit{ABSTRACT}}}
\label{sec:abstract}
\textbf{
 Most machine learning based regressors extract information from data  collected via past observations of limited length to make predictions in the future.  Consequently, when input to these trained models is data with significantly different statistical properties from data used for training, there is no guarantee of accurate prediction. Consequently, using these models on out of distribution input data may result in a completely different predicted outcome from the desired one, which is not only erroneous but can also be hazardous in some cases. Successful deployment of these machine learning models in any system requires a detection system, which should be able to distinguish between out-of-distribution  and in-distribution data (i.e. similar to training data). In this paper, we introduce a novel approach for this detection process using Reduced Robust Random  Cut Forest (RRRCF) data-structure, which can be used on both small and large data sets. Similarly to the Robust Random Cut Forest (RRCF), RRRCF is a structured, but reduced representation of the training data sub-space in form of cut-trees. Empirical results  of this method on both low and high dimensional data showed that inference about data being in/out of training distribution can be made efficiently and the model is easy to train with no difficult hyper-parameter tuning. The paper discusses  two different use-cases for testing and validating results. \url{https://github.com/vardhah/final_codes/tree/master/ODD}  
}
\subsection*{Keywords}
\textbf{\textit{ Random Cut Forest, Robust Random Cut Forest,interpretable intelligence,
Out-of-Distribution detection, Machine Learning,CARLA, Cyber Physical system}}
\section{Introduction}
\setlength{\parskip}{0pt}
\setlength{\parsep}{0pt}
\setlength{\headsep}{0pt}
\setlength{\topskip}{0pt}
\setlength{\topmargin}{0pt}
\setlength{\topsep}{0pt}
\setlength{\partopsep}{0pt}
\label{sec:introduction}

Machine Learning(ML) models are increasingly deployed in design and operation of engineering system\cite{vardhan2021machine}\cite{abbeel2010autonomous}, medical science\cite{al2019comparative}, financial sector \cite{ghazanfar2017using} etc. The training process of these models involve collection of data and then training on these collected data. During operation or prediction, we provide input to these trained ML models and get an output. As machine learning models are trained and validated against collected training data, their performance with new input  that is not consistent with the statistical properties  of the training data  cannot be relied on. Therefore, it is imperative to have some mechanism to verify that a given input data is  sampled from training data distribution. An Out-of-training-distribution (OOD) data point is one which is significantly different from the training data i.e. a data point which is an anomaly  relative to the training data so that it may stir speculation that it was generated by a different mechanism\cite{hawkins1980identification}.  


There are three major categories of  approaches to detect out of training distribution data: \textit{statistical detection techniques}, \textit{Deviation based techniques}, \textit{proximity based techniques} \cite{an2015variational}. 
Statistical detection techniques  attempt to fit the training data in a parametric/non-parametric probability distribution. The goal of learning is to find distribution model and parameters that can fit training data. Using this learned densities distribution, inference about new data point can be made based on probability of it being generated from the trained densities. A datapoint is defined as an OOD if the probability of it being generated is very low. The major limitation of this approach is difficulty to find a good probability fit, even when dimension of the problem increase beyond hundred.  
Deviation based approaches are based on one or other flavour of encoder-decoder setting. The encoder is trained to embed the data to latent space and  decoder is trained to reconstruct the data from latent space. During training of this model, goal is to reconstruct the input data as output of decoder. Once trained, inference about a datapoint can be made by passing it through encoder and decoder setting and estimating the reconstruction error/probability gap. The underlying assumption of deviation based methods are data from out-of-distribution will have high reconstruction error as the encoder-decoder parameters are not trained for it. 

Proximity based techniques assumes that anomalous data are isolated from the majority of the data and use various approaches to measure density or cluster representation and relationship of datapoint with the cluster. Our approach will fall into category of proximity based clustering approach for OOD detection. 
In clustering based approaches, an algorithm is deployed to represent the cluster of data from metric space to some  datastructure. Selection of algorithm and data-structure should be such that relationships of the data points in metric space must be preserved in this data-structure.The early work in this context is done by using randomised cut forests/isolation forests~\cite{liu2008isolation}. The isolation forest approach has several drawbacks, such as not compatible with  streaming data and missing crucial OODs in the presence of irrelevant dimensions etc\cite{guha2016robust}. To address these challenges, \cite{guha2016robust} proposed Robust Random Cut Forest(RRCF) as data structure, which is very promising in terms of very small false alarm rate (high accuracy), and can work on streaming data, but it is not scaleable on large data\cite{habeeb2019real}.  In this paper, we address this problem by developing a method that can use RRCF on large data sets. 
The  main idea of our approach is to make an  outline sketch of whole training data. This method capitalises on the fact that for sketching the data space, we only need few featured data point that can outline of data and can ignore all other irrelevant datapoint which has less/no relevance in creating the outline sketch. Here relevancy of a datapoint can be described as if inclusion of datapoint increase the dataspace coverage to a significant level then this datapoint is relevant for OOD detection. 
An elementary example is if two points are at same location in metric space then we can keep one as relevant datapoint and throw off other as irrelevant deta because inclusion of it will not contribute towards making decision about OOD.

The main contributions of this paper are the followings: 
\begin{enumerate}
    \item Developing a method which uses RRCF for OOD detection in computationally efficient manner on large data.  
    \item This proposed method is domain independent and can be used for any  ML-based regressor model. For empirical evaluation and validation of the proposed method, we tested it on two different test cases. First in Cyber Physical System (CPS) domain where we tested this method on a reinforcement learning controller with a three-dimensional data stream  Second, we tested this method on a high dimensional image data stream generated by an open source simulator CARLA \cite{dosovitskiy2017carla}.
    \item We conducted some empirical sensitivity analysis of this approach on the image data-set and found that this method can capture different level of changes in distribution.
    \item In contrast to deviation based method, which uses black box artificial neural network, The  proposed  method  follows  a  white-box  model  which is understandable  and  interpretable  (unlike neural networks) and can be used by the user to better understand the results. 
\end{enumerate}
 This approach has various benefits over other methods. First, it can make inference about the data being OOD/non-OOD on a stream of incoming data, which is compatible with the continuous operation of detection system. Second, empirical result reflects that it can capture presence of a single out-of-distribution data point in a stream of non-OOD data.
Third, for training this model there is no such difficult hyper-parameter tuning required. Hyper-parameter tuning in machine learning relies on experimental results, and general approach to determine the optimal settings is to try many different combinations or do random/guided search in hyper-parameter space and evaluate the performance of each model. This is an iterative process and can take plenty of training time. For training this model,We have only two hyper parameters. These are: \textit{number of trees} and \textit{size of tree}. Finding a right hyperparameter can be decided easily in comparison to other detectors(statistical/deviation based) which need to tune their hyper-parameter on various different aspects.  
Fourth, this approach is highly parallelizable so, by increasing allocation of resources the time complexity of the process can be improved. 



\section{Background}
\label{sec:discussion}
\setlength{\parskip}{0pt}
\setlength{\parsep}{0pt}
\setlength{\headsep}{0pt}
\setlength{\topskip}{0pt}
\setlength{\topmargin}{0pt}
\setlength{\topsep}{0pt}
\setlength{\partopsep}{0pt}
\captionsetup{justification=raggedright,singlelinecheck=false}
Our approach for anomaly detection relies on learning the data cluster (T) 's shape from metric space to a data-structure(S). The motivation behind learning the cluster's shape into a data-structure is to abstract the information from metric space in a structured manner such that computer and related algorithms can be efficiently deployed for inference.  If $D =\{d_1, d_2, ... , d_n\}$ are set of datapoints such that $d_i \in R^m $. For the purpose of Out-of-Distribution detection, following requirements are imposed on this data-structure(S):  
\begin{enumerate}
    \item $S$ should represent the cluster of data in structured way. 
    \item Relationships($\psi$) between the data points in metric space must be preserved in this data-structure. i.e 
    $$\psi\{T(d_k,d_l)\}\approx \psi\{S(d_k,d_l)\} $$
    \item Relationship($\phi$) of a data-point with the cluster can be encoded in simple quantitative measure. i.e. $\phi(T,d_k)$ can be measured as a scalar value in the data-structure($S$). 
 \end{enumerate}
 We chose the RRCF\cite{guha2016robust} as our data-structure to represent our cluster in structured way.  Robust Random Cut Forest can be formally defined as :
\begin{definition}
  Robust Random Cut Tree on set of data point $D=\{d_1,d_2,...,d_n\}$ can be generated by following procedure: 
 \end{definition}
 \begin{enumerate}
     \item $r_i= max_{X \in D}(X_i)-min_{X \in D}(X_i) \;\forall i\in m$
     \item $p_i= \dfrac{r_i}{\sum_{i=1}^{i=m}r_i} \;\forall i$
     \item select a random dimension $i$ with probability proportional to $p_i$
     \item $choose \; x_i \mid x_i \sim Uniform(max(X_i)-min(X_i))$
     \item $ D_1=\{X \mid X\in D ,X_i \leq x_i \}$
     \item $D_2= D \setminus D_1$
 \end{enumerate}
 $recurse\; on\; D_1\; and\; D_2\; until\; D_i \geq 1 $ 

Robust Random cut Forest is an ensemble of various RRCT. 
The selected relationship($\psi$) between datapoints in metric space is captured as $L_p$ distance between datapoints, then we require a distance preserving embedding of this relationship in the datastructure. For this purpose, the tree distance between two datapoints $d_k$ and $d_l$ in datastructure($S$) is defined as the weight of the least common ancestor of $d_k$ and $d_l$\cite{guha2016robust}, then according to Johnson-Landatrauss lemma\cite{lindenstrauss1984extensions} the tree distance can be bounded from atleast  $L_1 (d_k,d_l)$ to maximum $ O(d* log|k|/L_1(d_k,d_l))$. Accordingly, a point which is far from other points in metric space will continue to be at least as far in a random cut tree. 
Relationship($\phi$) of a data-point with the cluster can be encoded in simple quantitative measure by displacement which is an estimate of change in model complexity(summation of leave's depth) before and after inserting a given point $x$ in tree data-structure.

\section{Problem Formulation and Approach}
\setlength{\parskip}{0pt}
\setlength{\parsep}{0pt}
\setlength{\headsep}{0pt}
\setlength{\topskip}{0pt}
\setlength{\topmargin}{0pt}
\setlength{\topsep}{0pt}
\setlength{\partopsep}{0pt}

\setlength{\belowcaptionskip}{-10pt}
\captionsetup{justification=raggedright,singlelinecheck=false}

\label{sec:approach}
For problem formulation, we first introduce some notations. Let $X$ represents an input data point given to a machine learning model during training. This data can be an observation made by sensors attached to CPS system, like image data from camera or distance data from LIDAR etc. During training, we collect all such $X$s. From ML model's perspective, this collection of all observed $X$ represent environment in which model has been trained, we collectively call this set as $\mathbb{E}_t$.  
%
The OOD detection goal is to find whether input data $X'$ given during prediction is sampled from $\mathbb{E}_t$ or not. If $X' \sim \mathbb{E}_t$, then we call this observed state as non-OOD or else we call it as OOD, i.e. the trained agent has not seen this kind of input during training and the trained agent may behave unexpectedly to this OOD input data. 

Our approach to solve this problem is to learn the data cluster's shape from metric space to a Robust Random Cut Forest data-structure. The basic element of RRCF is a Robust Random Cut Tree(RRCT) that is a binary search tree constructed by recursively partitioning from given data points(Y) until each point is isolated. RRCF data structure contain sufficient information about the given data set(Y) and approximately preserve distances in metric space i.e. if a point is far from others that it will continue to be at least as far in a random cut tree in expectation and vice-versa (proof can be found in Guha et al \cite{guha2016robust}). Anomaly score(also called DispValue) of a data-point measures the change in model complexity incurred by inserting a given data-point $x$ in RRCF. Model complexity of a binary tree(RRCT) is defined as sum of depth of all datapoints in the tree. 
During insertion of a data-point that is far off the cluster in metric space , there is high probability to be partitioned in initial stage of RRCT construction. This will increase the depth of all leaf below it and consequently increase the model complexity by large number. On the other hand, if inserted point is inside the cluster, it will be partitioned in lower part of tree, and consequently it will have less number of leaf below it and this will reduce the model complexity by small amount. Given a training dataset $Y$, we first create these random cut trees and find the maximum Disp value of all inserted data-points. We define OOD as datapoint that is significantly different from data used in training i.e. it is  away from the training data cluster(Y) in the normed vector space. These off-the-cluster data-point has high probability to be isolated in initial stage of RRCT construction. Insertion of this point in tree will significantly increase the model complexity.

Applying this approach on large, high dimensional training data results in to creation of large forest and consequently any inference will have high prediction inaccuracy, computationally costly and even sometimes computationally infeasible. Underlying reason is as \textit{current learning models and processes are not very sample efficient}, and a well trained agent needs big training data. Research efforts are being made to make these process more sample efficient\cite{asadi2016sample} but it has few practical generality.\cite{genuer2017random} showed that on large data sub-sampling may improve random forest's performance but these forests are sensitive to extent of sub-sampling and become inconsistent with either no sub-sampling or too severe sub-sampling\cite{tang2018random}.  
\begin{algorithm}
 \SetAlgoLined
 \SetKwInOut{Input}{Input}
 \SetKwInOut{Output}{Output}
 \SetKwInOut{Parameter}{parameter}
 \caption{Reduced Robust Random Cut Forest(offline)}
 \Input{training data (Y)}
 \Output{reduced RRCF, threshold}
 \Parameter{number of trees}
 \label{alg:rrrcf}
 $Initialization: randomly\;select\; Z \mid  Z \subset Y $\\
 $rrcf_{init}= createForest(Z)$ \\
 $DispVal_{Z}= DispValue(Z)$\\
 $DispVal_{threshold}=mean(DispValue(Z))$\\
 $ L=Y \setminus Z$
 \\
 \For{$i = 0;\ i < len(L);\ i = i+1$}{
   \For{$j = 0;\ j <number of trees;\ j = i+1$}{
     $rrcf_{new}{(j)}=insertpoint(rrcf_{init}{(j)},L(i))$\\
     $mci(j)=Dispvalue(L(i))$ \\
     }
    $point_{dispValue}=mean(mci)$  \\
    \If{$point_{dispValue} \geq DispVal_{threshold}$}{
        \#include L(i) in featured Datapoint set \; 
        $DispVal_{Z}.append(point_{dispValue})$\\
        $DispVal_{threshold}=mean(DispValue(Z))$ \;
        
    }\Else{
     \#Do not include L(i) in featured Datapoint set \; 
      \For{$j = 0;\ j <number of trees;\ j = i+1$}{
         $rrcf_{new}{(j)}=Deletepoint(rrcf_{init}{(j)},L(i))$\\
        }
   }
  }
threshold $= max(DispVal_{Z})$ \\
return $rrcf_{new}$ , threshold \\
\end{algorithm}

To address these issue, we attempted to find only those featured datapoints which will create an outline of data cluster space and has significance in making decision about OOD and drop all others datapoints for construction of RRCF. The underlying intuition is in a dense data space, we need few datapoints to gain information about the data space they represent and we can drop most of the other datapoints. However in sparser region, we select most of the datapoint to represent the data space they acquire. The sparsity and density in dataspace is correlated with the average model complexity increment by inclusion of datapoint in data-structure. Using this approach, we can reduce the number of training data-points significantly that needs to be stored for construction of RRCF. RRCF created using these featured data-points is called Reduced Robust Random Cut Forest(RRRCF). For reduction of entire data to the featured data, we run a process of insertion and conditional deletion on each data point in training data. After one sweep of this process on whole training data, we collect all featured datapoints which represent identical data subspace as training data space. For initialisation, we first create RRCF using very small dataset(Z), where $Z \subset Y$, this will give us an initial small forest. Once we have created the RRCF using Z, we calculate the DispValue of all points in Z. For making decision on whether a given point can be included in the RRCF or not, we choose the mean of all DispValue calculated over Z as threshold. This threshold represents average complexity of datastructure. For rest of the datapoints ($Y \setminus Z$), we insert each point in forest and we calculate the DispValue for this point. If DispValue is more than the threshold DispValue of initial forest then we will keep this point in the forest or else we will reject this point and delete it from the RRCF. We recursively apply this on all left-over datapoints( $Y \setminus Z$). The final forest created from this process would be our reduced robust random cut forest and can be used for making inference during prediction for making decision about OOD for a given datapoint. This is an offline method and need to run only once. We can store the reduced RRCF(RRRCF) and threshold,  which is maximum model complexity from already included featured training datapoints of selected featured datapoints to be used for prediction(refer algorithm \ref{alg:rrrcf}).  This reduced forest is a structured representation of our training data sub-space.

\begin{algorithm}
\SetAlgoLined
\SetKwInOut{Input}{Input}
\SetKwInOut{Output}{Output}
\SetKwInOut{Parameter}{parameter}
\caption{OOD detector(online)}
\label{alg:ood}
\Input{RRRCF(T) , Data-point(x), threshold }
\Output{Inference about x being OOD}
 
 $T'=Insertpoint(T,x)$ \\
 $mci=calculate\, model\, complexity$ \\
 $T=Deletepoint(T',x) $\\
 $mcd=calculate\, model\, complexity$ \\
 $Disp Value =mci-mcd$ \\
 \uIf{$DispValue \geq threshold$}{
    x is OOD \;
  }
   \Else{
    x is not OOD \;
  }
\end{algorithm}

\begin{figure}[h!]
        \centering
        \captionsetup{justification=centering}
        \includegraphics[width=0.4\textwidth]{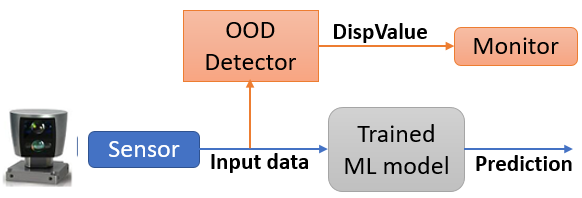}
        \caption{Deployment of OOD detector}
        \label{fig:detect_model}
    \end{figure}

During prediction, we insert newly observed input data point(x) into the stored RRRCF model(obtained from algorithm \ref{alg:rrrcf}) and check whether inclusion of this point increases the model complexity to an extent higher than threshold value.
Every new observation can be passed to this detector and prediction made by the machine learning based model can be accepted only if the scalar measure of datapoint by OOD detector is lower than the threshold generated by the algorithm \ref{alg:rrrcf}.  The setup for deploying this OOD detector is shown in fig \ref{fig:detect_model}.
     
If DispValue of new point is greater than the threshold obtained from algorithm \ref{alg:rrrcf}  then we declare this datapoint as OOD and viceversa (refer algorithm \ref{alg:ood}). For making inference on a datapoint,we do insertion step, where we include new datapoint to RRRCF and measure increased complexity and then we do deletion step to remove the inserted point. So,  during prediction, we do not extend our forest, we just do one insertion and one deletion step per prediction and our reduced robust random cut forest remain intact.

\section{Experiments}
\setlength{\parskip}{0pt}
\setlength{\parsep}{0pt}
\setlength{\headsep}{0pt}
\setlength{\topskip}{0pt}
\setlength{\topmargin}{0pt}
\setlength{\topsep}{0pt}
\setlength{\partopsep}{0pt}
\captionsetup{justification=raggedright,singlelinecheck=false}

\label{sec:experimentalResults}
To empirically evaluate above mentioned approach, we set up two experiments. In both cases, the machine learning model is a reinforcement learning based controller for car braking system where in first experiment, observations are in low dimension( 3 dimensions) while in second experiment, our observation is in high dimension image data. 

\begin{figure}[h!]
        \centering
        \captionsetup{justification=centering}
        \includegraphics[width=0.45\textwidth]{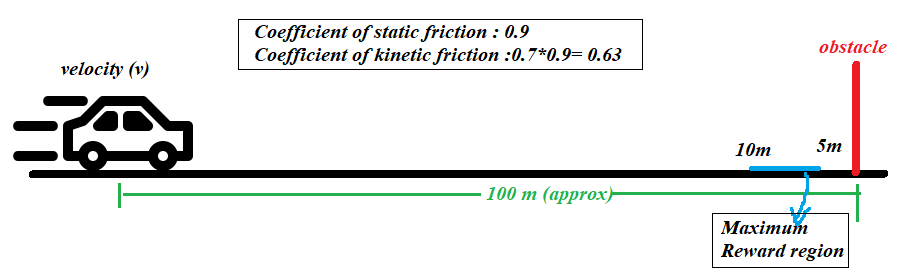}
        \caption{Training Scenario setup}
        \label{fig:exp1}
    \end{figure}

In the \textit{first experiment}, an approaching car detects a \textbf{stationary obstacle} at distance of 100 meters and the learning goal is to self-train a controller for braking system to stop the car without crashing (refer fig \ref{fig:exp1}). We used reinforcement learning algorithm called DDPG (Deep Deterministic Policy Gradient)\cite{lillicrap2015continuous} based learning model to design the braking system of a car. The static and kinetic friction coefficient of road is constant throughout the experiment. The random variable in this scenario is the speed of the car when it detects the obstacle, which is drawn from a uniform distribution between 40 to 70 miles/hour.  $Initial\_Speed(v) \sim \mathcal{U}(40,70)$. The reward setting is done in such a way that vehicle when brakes around region of 5-10 meters from obstacle, gets the maximum reward.  
During training, we observe three variables, $d$ (distance form obstacle), $v$ (velocity of car) and $mu$ (friction coefficient). $X=\{d,v,mu\}$ provides state information of vehicle in the environment. During training, $X$ becomes the input of the neural network and brake value $b$ is the output. 
    
\begin{figure}[h!]
        \centering
        \captionsetup{justification=centering}
        \includegraphics[width=0.45\textwidth]{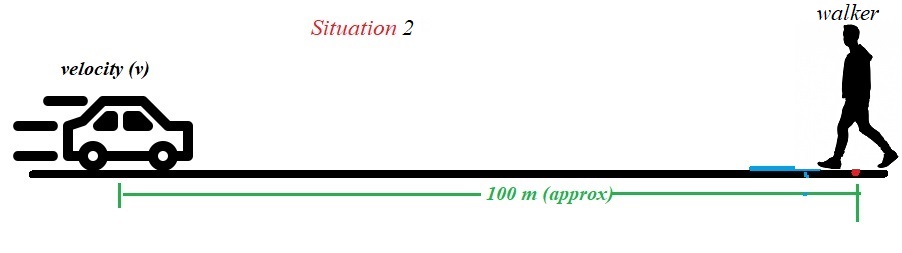}
        \caption{Prediction Scenario I}
        \label{fig:exp1_cs}
    \end{figure}

The braking system is trained and tested in this environmental setting. During \textit{prediction}, we make some changes in the environment, which was never observed during training process. We created two such scenarios: first, in place of stationary obstacle we used an obstacle that is moving toward the car at some small velocity(for this experiment it is drawn from a uniform random distribution between 0.1-2 meter/second, refer figure \ref{fig:exp1_cs}, moving obstacle is represented by a walker). This situation was never observed during training, and result of it the braking system do not respond to this changed scenario appropriately and it leads to crash. 
 In the second scenario, we keep obstacle stationary but spawned the vehicle at the velocity 75 miles/hr. It means the vehicle when detect the obstacle and invokes the braking system has velocity out of the training range (40-70 miles/hr). In this scenario also, the braking system fails to brake and leads to to a crash because during training we never observed this speed.
  \begin{figure}[h!]
        \centering
        \captionsetup{justification=centering}
        \includegraphics[width=0.48\textwidth]{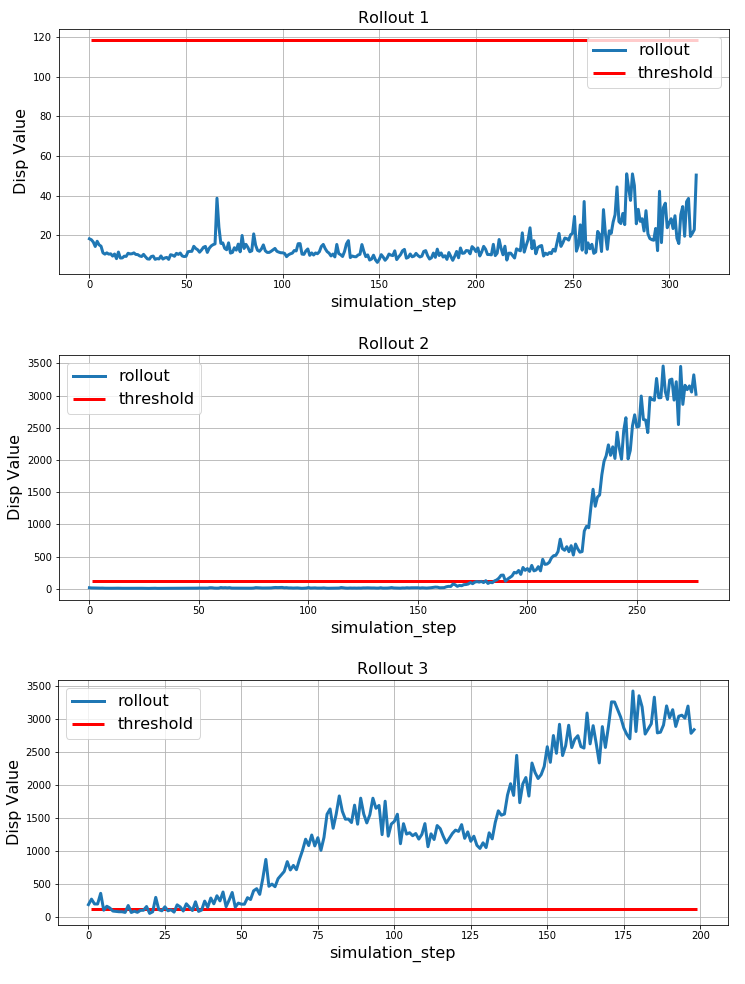}
        \caption{Three rollouts and respective output Dispvalue and threshold . Rollout1: Environment is same as training scenario(stationary obstacle with initial speed $\sim \mathcal{U}$(40,70)) ; Rollout2: Moving obstacle(walker) with initial speed $\sim \mathcal{U}$(40,70) ; Rollout3: Stationary obstacle with initial speed(v) $\nsim \mathcal{U}$(40,70) }
        \label{fig:exp1_results}
    \end{figure}

In both cases, our RRRCF based detection scheme detect these evolved situations and raised a flag for data being out of distribution. Figure \ref{fig:exp1_results} shows result of above three different test roll-outs.
\textit{Dispvalue} which is a scalar measure for each datapoint given to this RRRCF detector provides inference  about  possibility  of  data  being  in  or  out  of training  distribution. The threshold maximum \textit{Dispvalue} derived for this set of experiment by running algorithm \ref{alg:rrrcf} is approximately $118.58$, which is shown by the red horizontal line in figure \ref{fig:exp1_results}.
During test rollouts, at each step of simulation, we estimated \textit{DispValue} for datapoint observed i.e. $X=\{d,v,mu\}$ as simulation progress. 
 
 In the \textit{first} roll-out(Rollout 1), we did not make any change in the environment and environment was similar to the training scenarios. The observed \textit{Dispvalue} for each datapoint during rollout is always less than the threshold value, which implies there was not any state observed during this rollout which is OOD. In case of second roll-out(Rollout2), we made the obstacle move towards the car at the speed of 2 meter/second. It was observed that in the initial stage of simulation, the \textit{Disp value} is less than threshold value. It is because as obstcale is moving slowly, at initial stage very small change in observed state. Once car approaches the obstacle, the new state information $X$ is slowly shifting from (i.e. getting far off) from training data cluster resulting into higher and higher \textit{Disp value} as car reaches towards the end of rollout. It is also observed that, the change is Disp value is quite significant which facilitate easy detection of such scenario and less false alarms. In case of third roll-out(Rollout3), we keep the obstacle stationary but we changed the initial spawning velocity of car to 75 miles/hour. We observed \textit{Dispvalue} higher than threshold almost through-out the simulation step, the underlying reason initial spawned velocity of car is out of training range and when OOD detector observe this first initial state , it finds it as out-of-distribution. It can be also seen that initially measured Dispvalue fluctuates around the threshold, which reflects very proximal OODs but with the progress of simulation, this gap increases and results into higher value of measured DispValue. 

 \begin{figure}[h!]
        \centering
        \captionsetup{justification=centering}
        \includegraphics[width=0.48\textwidth]{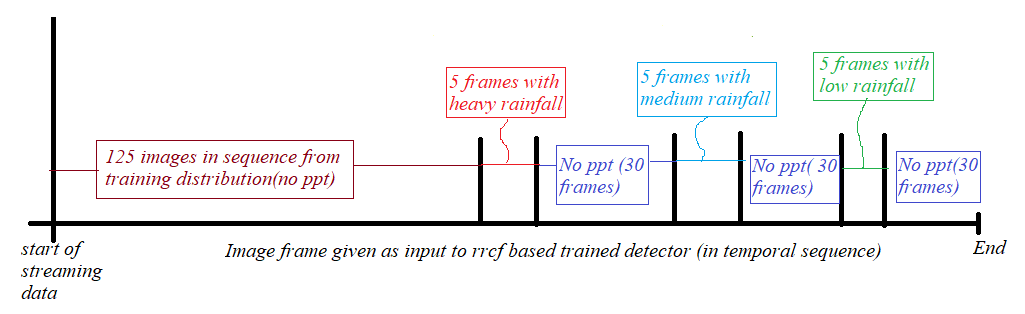}
        \caption{Image data stream given as an input to OOD detector (Stream I)}
        \label{fig:ids_5}
    \end{figure}

\begin{figure}[h!]
        \centering
        \captionsetup{justification=centering}
        \includegraphics[width=0.48\textwidth]{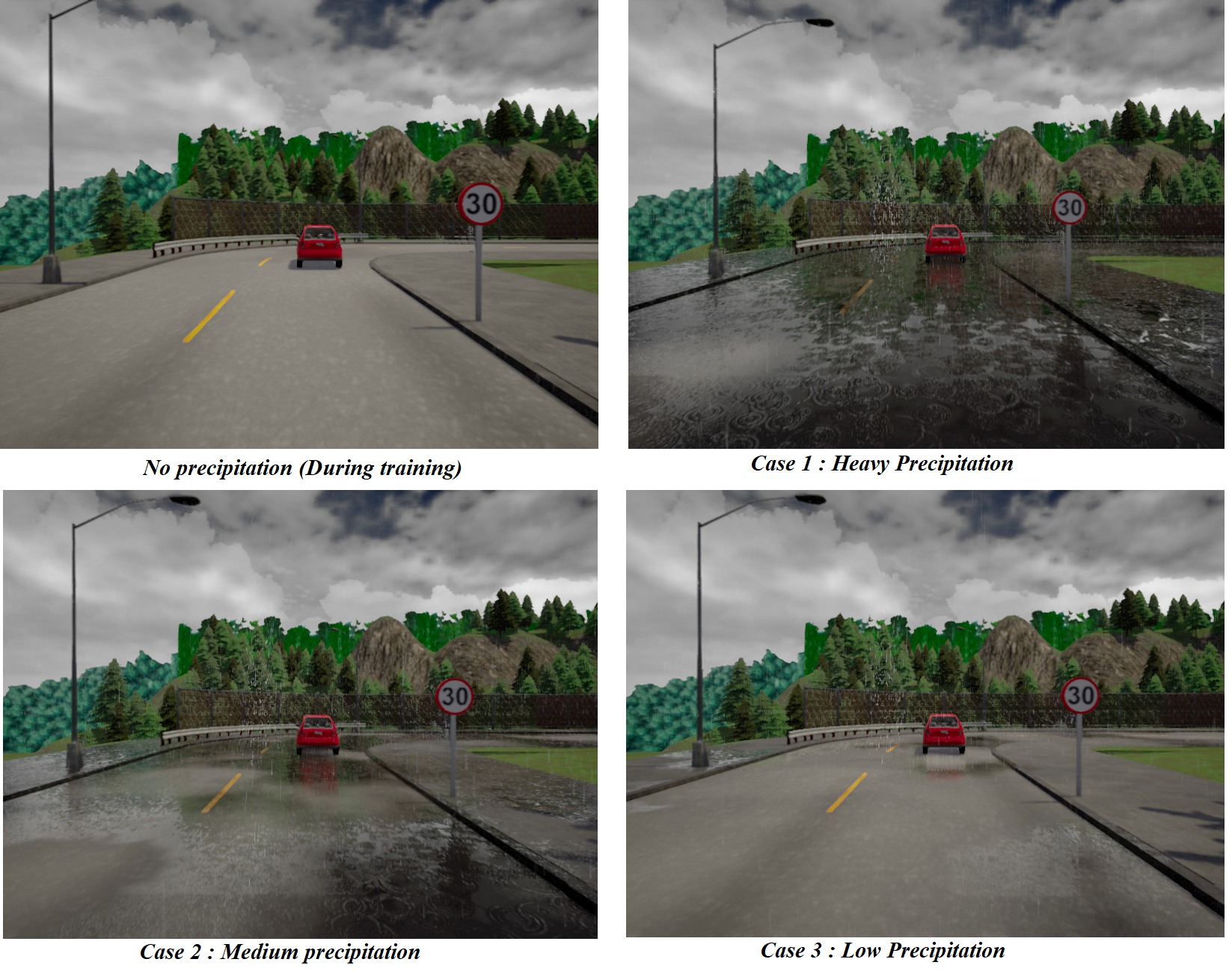}
        \caption{Different precipitation conditions generated during the training and prediction \textit{(no ppt, heavy ppt, medium ppt, low ppt)}}
        \label{fig:allppt}
    \end{figure}

In the \textit{second} experiment, we used the work done by Cai et al \cite{cai2020real} for training a braking controller, but we trained it on wider range of speed(30-75 m/s) with different reward policy. Input in this case is the image scene from the car's camera and predicted output is the brake value. We trained and tested this model on image data generated with \textbf{no precipitation} condition in simulator CARLA. We collected all these images with no precipitation in several episodic rollout and label it as non-OOD training scenario data. We used all these image data to build our RRRCF based OOD detector by running algorithm \ref{alg:rrrcf}. Threshold scalar \textit{Disp Value} in this experiment is calculated and it has a value of $23.26$.
\begin{figure}[h!]
        \centering
        \captionsetup{justification=centering}
        \includegraphics[width=0.48\textwidth]{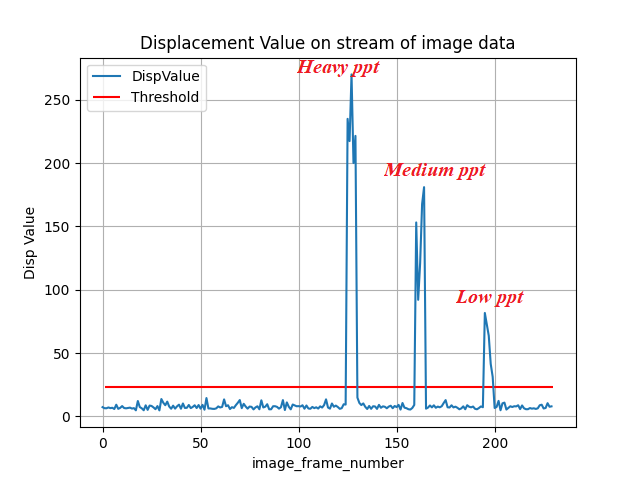}
        \caption{Output \textit{Disp value} of stream I }
        \label{fig:exp2_5_result}
    \end{figure}   
For creating OOD scenarios, we changed the no-precipitation condition  to three different participation conditions, \textit{Case1:} Heavy precipitation, \textit{Case2:} Medium precipitation, and \textit{Case3:} Low precipitation. (refer fig \ref{fig:allppt} ) and collected image data captured by the camera.

Our goal is to evaluate the performance of our OOD detector on stream of image data. For this purpose we created a stream of images called \textbf{stream I},which has first 125 image frames with no precipitation scenario, then $5$ frames of each three different precipitation cases(heavy, medium and low) with 30 intermittent frames of no precipitation in-between as shown in figure \ref{fig:ids_5}. 
\begin{figure}[h!]
        \centering
        \captionsetup{justification=centering}
        \includegraphics[width=0.48\textwidth]{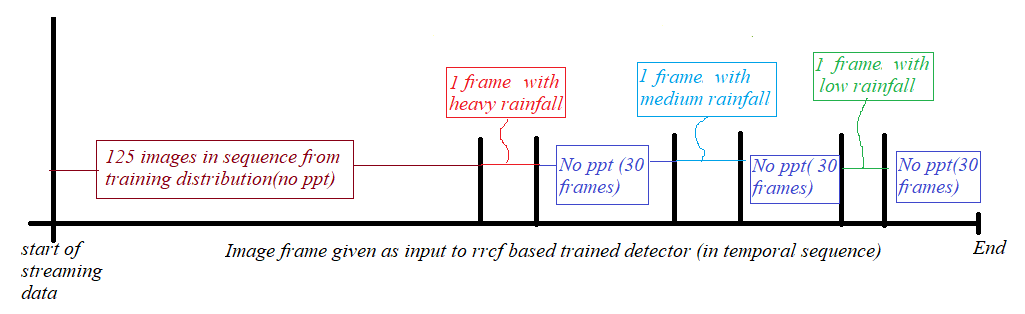}
        \caption{Image data stream given as input to OOD detector (Stream II)}
        \label{fig:ids_1}
    \end{figure}
 
\begin{figure}[h!]
        \centering
        \captionsetup{justification=centering}
        \includegraphics[width=0.48\textwidth]{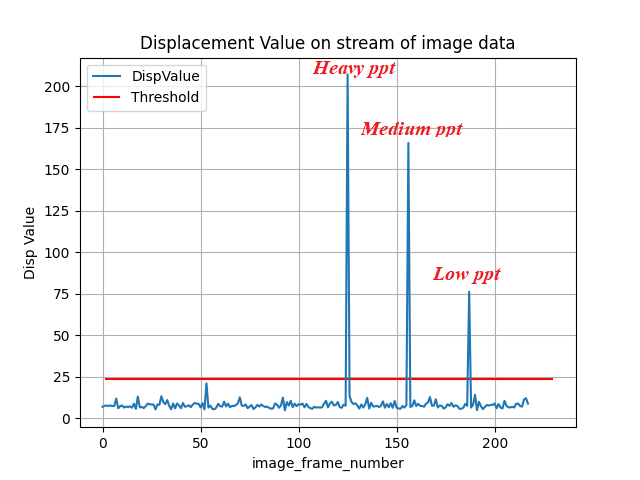}
        \caption{Output \textit{Disp value} of stream II }
        \label{fig:exp2_1_result}
    \end{figure}      

 With a given input image, if \textit{Disp Value} is greater than the threshold value then we declare it as OOD. Figure \ref{fig:exp2_5_result} shows the \textbf{Disp value} generated by the OOD detector for the data stream I. The red horizonal line represent the threshold value of detector. It can be easily observed that the detector output is significantly higher than threshold value in all the three OOD case(heavy, medium and low precipitation)  and lower for all non-OOD cases. 

we tried to find the performance of detector in context of point OOD i.e. is the detector able to detect even single OOD in stream of non-OOD data?
 
To find an answer to this question, we created another stream of image frames called, \textbf{Stream II}. In Stream II, we inserted single frame of different precipitation cases(heavy, medium and low) in a stream of image frame similar to training data i.e. no precipitation (refer \ref{fig:ids_1}). We observed that OOD detector can even detect a single OOD data in a stream of no-OOD data (for result,refer figure \ref{fig:exp2_1_result}).

In second experiment, our model is a design predictor\cite{vardhan2021machine} which predict geometric design (chord profile radial distribution,diameter of propeller fins, hub diameter) and efficiency on a given requirement. Due to with infinitely large search space, it is not possible to explore entire design space and these models have knowledge only about the part of design space\cite{vardhan2021machine}. The knowledge of design sub-space in which prediction of these models can be relied is important for reliability of these models. The detection of inputs which are out-of-distribution can flag to inputs which are not in our model's capability and prediction on these input can not be relied. On sampled 10 million design points and found 0.205 million valid design points.   
\\

\section{Discussion}
\label{sec:discussion}
\setlength{\parskip}{0pt}
\setlength{\parsep}{0pt}
\setlength{\headsep}{0pt}
\setlength{\topskip}{0pt}
\setlength{\topmargin}{0pt}
\setlength{\topsep}{0pt}
\setlength{\partopsep}{0pt}
\captionsetup{justification=raggedright,singlelinecheck=false}
\subsection{Time Complexity Analysis}
\begin{figure}[h!]
        \centering
        \captionsetup{justification=centering}
        \includegraphics[width=0.4\textwidth]{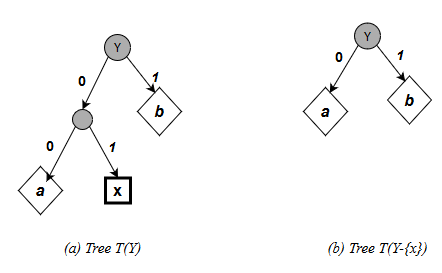}
        \caption{Deletion of a data point from a tree}
        \label{fig:tree_deletion}
    \end{figure}

Algorithmic complexity of ODD detection process should be low during prediction time. Its time complexity depends upon the process required to make inference on a data-point by detector i.e. "Forest maintenance on stream". Forest Maintenance on stream of data involves two processes per data-point first, \textit{insertion of a point in tree data-structure} and second, \textit{deletion of a point from tree data-structure}. 
If we store each data-point i.e the leaf of a tree in a hash table, then search process for locating a leaf will have time complexity $O(1)$. Given a set of points $Y$ and a point $x \in Y$ , we construct a tree $T$ on data $Y$. Consider if we want to delete a leaf $x$ from the tree and produces tree $T(Y - x$), we just need to remove the parent of $x$ and make another child's parent pointer to its grandparent accordingly (refer fig \ref{fig:tree_deletion}). This deletion process has time complexity of $O(1)$. Total time complexity of deletion process of a point (search+deletion) from a tree is $O(1)$.

We can efficiently insert and delete data-points into a random cut tree data-structure. 
    
    Insertion of a point in tree is a process when we have a tree $T$ and we want to insert a leaf $x$ to the tree and produces tree $T(Y \cup x)$, where $x \notin Y$. In the worst case, the time complexity of this process would be O(n), while in the average and best case it would be O(logn) and O(1) respectively, where n is the size of tree i.e. number of data-points used to create the tree. If we have $m$ trees in a forest, then total time complexity of the process is O(mn), O(mlogn) and O(m) in worst, average and best case respectively. The effect of $m$ on RRRCF creation,insertion and deletion process can be made O(1) for time complexity analysis purpose as the process is completely parallelizable i.e. we can do insertion and deletion on each tree on different thread/core in parallel if GPU/multi-core based parallel implementation of RRRCF is used. Consequently, the insertion process on forest will have time complexity O(n), O(logn) and O(1) respectively for worst, average and best case scenario. 
    Here a reasonable question is how big the variable $n$ can be? As machine learning process is not very sample efficient, a well trained model results into large training data. For example, training a braking system perception LEC for speed range variance of 10Km/hr(90km/h to 100km/h) in Cai et al \cite{cai2020real} experimental setup took approximately 8160 images in a 100 meter distance. Similarly in experiment set 1, our training data-point was approx 0.36 million. But for ODD detection, it is not required to use whole training data  and it is enough if we can is to make a sketch/outline of data i.e. we can ignore many datapoints that are densly clustered and use only those featured datapoints which represent the outline of data subspace.

 We used our offline algorithm to select the number of data points which may represent the whole data. For first set of experiment,our algorithm gave 11134 featured data-points out of total 367035 datapoints. For second set of experiment,our algorithm gave 1050 featured data-points out of total 8197 image datapoints. Selection on $n$ also depends upon the resources available. With availability of large number of parallel cores in modern GPU, it is also possible to increase $m$ and reduce $n$ and randomly choose a subset of element out of $n$ for creating a tree. With a very large training data, it is possible to extract only a small segment of data to sketch the RRRCF outline.

\subsection{Sensitivity of OOD detector on change in distribution:}  A good OOD detector should be able to detect  even a very small deviation in input data from training data distribution. For testing the sensitivity of the OOD detector, in experiment set 2, we used image frames generated from different weather condition (different level of precipitation: heavy, medium, and low. refer figure \ref{fig:allppt}). Higher the precipitation level, the more deviation of data from training distribution.  Empirical result suggests that this RRRCF based detector can detect all three different level of change in distribution. Empirical result also suggests that as test data-point go away from the training data cluster in  metric space, the  anomaly score i.e. DispValue progressively increases in expectation.

\subsection{Sensitivity of OOD detector on single OOD data in stream of non-OOD data:}  Most work in the field of  out-of-distribution has been done in the context of classification tasks in computer vision. OOD detection problem in machine learning can generally classified in two parts- contextual OOD and point OOD. In case of contextual OOD, goal is to detect the anomaly in observed input data in environmental context.  SThere are two  These detection techniques do not take in to consideration of dynamic nature of CPS and will not be suitable to use in it. \cite{cai2020real} proposed a method for OOD detection for CPS system but this method needs a number of OOD data before making conclusion about the OOD. It may be fail in scenarios where we do not have continuous stream of OOD data. rrrcf has tried to fill this gap, where it can detect even a single OOD data in the stream of non-OOD data. 


\subsection{Ease of training:} Other OOD detectors like GMM based and VAE based need to be tuned by selecting various hyper-parameters. These hyper-parameter tuning and re-training is an iterative process and search of an effective hyperparameter is slow process and time consuming. In contrast, \textbf{\textit{rrrcf}} has only two hyperparameter- first, \textit{number of trees in forest} and second, \textit{depth of each tree}. As rrrcf works on expectation value of anomaly score, any reasonable number of trees would work. We used 100 trees for both experimental setup. Second hyperparameter is actually derived from our algorithm \ref{alg:rrrcf} so, we do not need to tune it anyway while creating a rrrcf.  

\section{Training and Experiment Details}
\setlength{\parskip}{0pt}
\setlength{\parsep}{0pt}
\setlength{\headsep}{0pt}
\setlength{\topskip}{0pt}
\setlength{\topmargin}{0pt}
\setlength{\topsep}{0pt}
\setlength{\partopsep}{0pt}
\captionsetup{justification=raggedright,singlelinecheck=false}

\label{sec:ledetails}

\subsection{RRRCF Training Details :}
In case of experiment set 1, we collected approximately 367035 training datapoints(Y), where each data point is a tuple of distance, velocity and friction coefficient(\{d,v,mu\}). For constructing initial RRCF, we selected 1000 random data from pool of total data i.e Z=1000. After construction of initial rrcf datastructure , mean DispValue is calculated over all points in Z. It is the initial threshold value and it's value was approximately $6.4$ . This threshold was used for making decision about inclusion/discard of other points in training data ($Y \setminus Z$). After every 5000 data-points, we recalculated the mean DispValue as our new threshold. The reason for repeatedly updating our threshold is to accomodate the changes made in tree structure  by inclusion of datapoints. We recursively applied this process to rest of training data. 
After running algorithm \ref{alg:rrrcf}, 10134 featured datapoints were selected and  rest 355574 datapoints were rejected. These total 11134 (10134+1000) datapoints were used to represent our whole data(367035) in the reduced RRCF datastructure, which is approximately $3.03 \%$ percent of whole data. 
\begin{table}
\captionsetup{justification=centering,labelsep=period}
\caption{Total datapoints and featured datapoints for reduced robust random cut and percentage of total datapoints selected as featured datapoint }
\label{tab:reduction}
\begin{tabular}{c|c|c|c} 
\hline
\textbf{Scenario} & \textbf{Training points} & \textbf{Featured points} &\textbf{\% reduction} \\
\hline
1 & 367035 & 11134  & 96.97 \\
\hline
2 & 8197 & 1050 & 87.2  \\
\hline

\end{tabular}
\end{table}
In case of experiment set 2, same experimental setting as by Cai et al \cite{cai2020real} was used and 8197 images were collected for training scenario i.e. with no precipitation condition. For creation of initial RRCF, we selected 100 images i.e Z=100. Once initial RRCF was created, the average threshold for Z was calculated and it was approximately $5.62$. This threshold was used for making decision about inclusion/discard of other points in training data ($Y \setminus Z$). After every 100 data-points, we recalculated the mean DispValue as our new threshold. After running algorithm \ref{alg:rrrcf}, 950 datapoints were selected and rest 7147 datapoints were rejected. These featured 1050 datapoints(950+100) were used to represent our whole data(8197 images) in the reduced RRCF datastructure, which is approximately $12.8\%$ percent of whole data. For construction of random cut tree, modified version of implementation of tree written by Barto et al \cite{bartos2019rrcf} was used. 

\subsection{Braking system Training Details:} The braking system in experiment set 1 is trained using Actor-Critic based Deep Deterministic Policy Gradient algorithm with both actor and critic network are three layer neural network, with 50(layer1) and 30(layer2) neurons in the hidden layer and 1 neuron in output layer. For hidden layers in both cases, relu activation function was used and for output layer sigmoid activation layer was used in actor and linear activation was used in critic. The model was trained using Adam\cite{kingma2014adam} optimization method by tuning different learning rate for 5000 episodes.  In case of experiment set 2, we used same setting  as done by Cai et al \cite{cai2020real}.
\section{Related Work}
\setlength{\parskip}{0pt}
\setlength{\parsep}{0pt}
\setlength{\headsep}{0pt}
\setlength{\topskip}{0pt}
\setlength{\topmargin}{0pt}
\setlength{\topsep}{0pt}
\setlength{\partopsep}{0pt}

\label{sec:relatedWorks}
It is a well known issue that machine learning models fail when the
training and test distributions differ and they often do this even after providing high confidence predictions on training data \cite{amodei2016concrete}. 
Out of distribution detection can be seen as attempt towards making a verified and safe artificial intelligence. As direct verification suffer from a scalability problem due to computational complexity and size of networks, till now work has been done for smaller scale networks or with approximate methods that provide some convergence guarantees on the bounds\cite{huang2020survey}. The alternative is to detect data points those are OOD and do not use predicted output from the trained model on these kind of input data.
\cite{hendrycks2016baseline} attempted the detect OOD based on the softmax probability estimated by the predictor. Empirically they showed that OOD data generally have lower softmax probabilities than the correctly classified datapoints, as this can be used to detect the data being OOD.  
\cite{an2015variational}\cite{cai2020real}\cite{sundar2020out} used the deviation based approach using one or other flavour of Variational Auto Encoder(VAE). VAEs are directed probabilistic graphical model, whose posterior and likelihood are approximated by a neural networks,  forming an encoder and decoder like structure. 
In VAE based OOD detector, people capitalise on either latent space representation or its error in reconstruction to find the OOD data from normal data. 
In proximity based method, early work on OOD detection is done by using collection of random cut decision tree and called it isolation forest\cite{liu2008isolation}. Improving this work \cite{guha2016robust} proposed Robust Random cut forest , which addressed most of the challenge posed during practical use of isolation forest like: working on stream of data, false detection etc.

\section{Conclusion and Future Work}
\setlength{\parskip}{0pt}
\setlength{\parsep}{0pt}
\setlength{\headsep}{0pt}
\setlength{\topskip}{0pt}
\setlength{\topmargin}{0pt}
\setlength{\topsep}{0pt}
\setlength{\partopsep}{0pt}
\captionsetup{justification=raggedright,singlelinecheck=false}
\label{sec:conclusionFutureWork}

In this work, we demonstrated a white-box interpretable method for out-of-distribution detection. 
We also showed that this method can detect even a single OOD data in a stream of non-OOD data. We demonstrated the effectiveness of this approach for both low and high dimensional input data space. 

We also discussed that a GPU based parallel implementation of reduced RRCF can significantly reduce the execution time. Parallel implementation of reduced rrcf may be used in real time system for real-time OOD detection.  
Writing an open source GPU and multi-core implementation of reduced RRCF and its evaluation for real time performance is the part of future work. Evaluation on real world data and comparison with other methods for OOD like variational autoencoder (VAE) based, Gaussian mixture models based  are also part of our future work.

\section{Acknowledgments}
This work is supported by DARPA’s Symbiotic Design for CPS project and by the Air Force Research Laboratory (FA8750-20-C-0537).
Any opinions, findings, and conclusions, or recommendations expressed in this material are those of the author(s) and do not
necessarily reflect the views of DARPA or AFRL.\cite{10.1145/3468891.3468897}

\bibliographystyle{IEEEtran}
\bibliography{references}

\begin{thebibliography}{10}
\providecommand{\url}[1]{#1}
\csname url@samestyle\endcsname
\providecommand{\newblock}{\relax}
\providecommand{\bibinfo}[2]{#2}
\providecommand{\BIBentrySTDinterwordspacing}{\spaceskip=0pt\relax}
\providecommand{\BIBentryALTinterwordstretchfactor}{4}
\providecommand{\BIBentryALTinterwordspacing}{\spaceskip=\fontdimen2\font plus
\BIBentryALTinterwordstretchfactor\fontdimen3\font minus
  \fontdimen4\font\relax}
\providecommand{\BIBforeignlanguage}[2]{{%
\expandafter\ifx\csname l@#1\endcsname\relax
\typeout{** WARNING: IEEEtran.bst: No hyphenation pattern has been}%
\typeout{** loaded for the language `#1'. Using the pattern for}%
\typeout{** the default language instead.}%
\else
\language=\csname l@#1\endcsname
\fi
#2}}
\providecommand{\BIBdecl}{\relax}
\BIBdecl

\bibitem{vardhan2021machine}
H.~Vardhan, P.~Volgyesi, and J.~Sztipanovits, ``Machine learning assisted
  propeller design,'' in \emph{Proceedings of the ACM/IEEE 12th International
  Conference on Cyber-Physical Systems}, 2021, pp. 227--228.

\bibitem{abbeel2010autonomous}
P.~Abbeel, A.~Coates, and A.~Y. Ng, ``Autonomous helicopter aerobatics through
  apprenticeship learning,'' \emph{The International Journal of Robotics
  Research}, vol.~29, no.~13, pp. 1608--1639, 2010.

\bibitem{al2019comparative}
A.~Al~Bataineh, ``A comparative analysis of nonlinear machine learning
  algorithms for breast cancer detection,'' \emph{International Journal of
  Machine Learning and Computing}, vol.~9, no.~3, pp. 248--254, 2019.

\bibitem{ghazanfar2017using}
M.~A. Ghazanfar, S.~A. Alahmari, Y.~F. Aldhafiri, A.~Mustaqeem, M.~Maqsood, and
  M.~A. Azam, ``Using machine learning classifiers to predict stock exchange
  index,'' \emph{International Journal of Machine Learning and Computing},
  vol.~7, no.~2, pp. 24--29, 2017.

\bibitem{hawkins1980identification}
D.~M. Hawkins, \emph{Identification of outliers}.\hskip 1em plus 0.5em minus
  0.4em\relax Springer, 1980, vol.~11.

\bibitem{an2015variational}
J.~An and S.~Cho, ``Variational autoencoder based anomaly detection using
  reconstruction probability,'' \emph{Special Lecture on IE}, vol.~2, no.~1,
  pp. 1--18, 2015.

\bibitem{liu2008isolation}
F.~T. Liu, K.~M. Ting, and Z.-H. Zhou, ``Isolation forest,'' in \emph{2008
  eighth ieee international conference on data mining}.\hskip 1em plus 0.5em
  minus 0.4em\relax IEEE, 2008, pp. 413--422.

\bibitem{guha2016robust}
S.~Guha, N.~Mishra, G.~Roy, and O.~Schrijvers, ``Robust random cut forest based
  anomaly detection on streams,'' in \emph{International conference on machine
  learning}, 2016, pp. 2712--2721.

\bibitem{habeeb2019real}
R.~A.~A. Habeeb, F.~Nasaruddin, A.~Gani, I.~A.~T. Hashem, E.~Ahmed, and
  M.~Imran, ``Real-time big data processing for anomaly detection: A survey,''
  \emph{International Journal of Information Management}, vol.~45, pp.
  289--307, 2019.

\bibitem{dosovitskiy2017carla}
A.~Dosovitskiy, G.~Ros, F.~Codevilla, A.~Lopez, and V.~Koltun, ``Carla: An open
  urban driving simulator,'' \emph{arXiv preprint arXiv:1711.03938}, 2017.

\bibitem{lindenstrauss1984extensions}
W.~J.~J. Lindenstrauss, ``Extensions of lipschitz maps into a hilbert space,''
  \emph{Contemp. Math}, vol.~26, pp. 189--206, 1984.

\bibitem{asadi2016sample}
K.~Asadi and J.~D. Williams, ``Sample-efficient deep reinforcement learning for
  dialog control,'' \emph{arXiv preprint arXiv:1612.06000}, 2016.

\bibitem{genuer2017random}
R.~Genuer, J.-M. Poggi, C.~Tuleau-Malot, and N.~Villa-Vialaneix, ``Random
  forests for big data,'' \emph{Big Data Research}, vol.~9, pp. 28--46, 2017.

\bibitem{tang2018random}
C.~Tang, D.~Garreau, and U.~von Luxburg, ``When do random forests fail?'' in
  \emph{Advances in neural information processing systems}, 2018, pp.
  2983--2993.

\bibitem{lillicrap2015continuous}
T.~P. Lillicrap, J.~J. Hunt, A.~Pritzel, N.~Heess, T.~Erez, Y.~Tassa,
  D.~Silver, and D.~Wierstra, ``Continuous control with deep reinforcement
  learning,'' \emph{arXiv preprint arXiv:1509.02971}, 2015.

\bibitem{cai2020real}
F.~Cai and X.~Koutsoukos, ``Real-time out-of-distribution detection in
  learning-enabled cyber-physical systems,'' in \emph{2020 ACM/IEEE 11th
  International Conference on Cyber-Physical Systems (ICCPS)}.\hskip 1em plus
  0.5em minus 0.4em\relax IEEE, 2020, pp. 174--183.

\bibitem{bartos2019rrcf}
M.~D. Bartos, A.~Mullapudi, and S.~C. Troutman, ``rrcf: Implementation of the
  robust random cut forest algorithm for anomaly detection on streams,''
  \emph{Journal of Open Source Software}, vol.~4, no.~35, p. 1336, 2019.

\bibitem{kingma2014adam}
D.~P. Kingma and J.~Ba, ``Adam: A method for stochastic optimization,''
  \emph{arXiv preprint arXiv:1412.6980}, 2014.

\bibitem{amodei2016concrete}
D.~Amodei, C.~Olah, J.~Steinhardt, P.~Christiano, J.~Schulman, and D.~Man{\'e},
  ``Concrete problems in ai safety,'' \emph{arXiv preprint arXiv:1606.06565},
  2016.

\bibitem{huang2020survey}
X.~Huang, D.~Kroening, W.~Ruan, J.~Sharp, Y.~Sun, E.~Thamo, M.~Wu, and X.~Yi,
  ``A survey of safety and trustworthiness of deep neural networks:
  Verification, testing, adversarial attack and defence, and
  interpretability,'' \emph{Computer Science Review}, vol.~37, p. 100270, 2020.

\bibitem{hendrycks2016baseline}
D.~Hendrycks and K.~Gimpel, ``A baseline for detecting misclassified and
  out-of-distribution examples in neural networks,'' \emph{arXiv preprint
  arXiv:1610.02136}, 2016.

\bibitem{sundar2020out}
V.~K. Sundar, S.~Ramakrishna, Z.~Rahiminasab, A.~Easwaran, and A.~Dubey,
  ``Out-of-distribution detection in multi-label datasets using latent space of
  beta-vae,'' \emph{arXiv preprint arXiv:2003.08740}, 2020.

\bibitem{10.1145/3468891.3468897}
\BIBentryALTinterwordspacing
H.~Vardhan and J.~Sztipanovits, ``Rare event failure test case generation in
  learning-enabled-controllers,'' in \emph{2021 6th International Conference on
  Machine Learning Technologies}, ser. ICMLT 2021.\hskip 1em plus 0.5em minus
  0.4em\relax New York, NY, USA: Association for Computing Machinery, 2021, p.
  34–40. [Online]. Available: \url{https://doi.org/10.1145/3468891.3468897}
\BIBentrySTDinterwordspacing

\end{thebibliography}

\end{document}